%% file: PaperForReview.tex
\documentclass[10pt,twocolumn,letterpaper]{article}

\usepackage{cvpr}              %

\usepackage{graphicx}
\usepackage{amsmath}
\usepackage{amssymb}
\usepackage{booktabs}
\usepackage{mathtools}

\usepackage[pagebackref,breaklinks,colorlinks]{hyperref}
\DeclarePairedDelimiter{\abs}{\lvert}{\rvert}

\usepackage[capitalize]{cleveref}
\crefname{section}{Sec.}{Secs.}
\Crefname{section}{Section}{Sections}
\Crefname{table}{Table}{Tables}
\crefname{table}{Tab.}{Tabs.}

\def\cvprPaperID{*****} %
\def\confName{CVPR}
\def\confYear{2022}

\begin{document}

\title{2D Pre-Training for 3D Pose Estimation}

\author{Liyao Jiang\\
University of Alberta\\
{\tt\small liyao1@ualberta.ca}
\and
Ruichen Chen\\
University of Alberta\\
{\tt\small ruichen1@ualberta.ca}
\and
Keith G. Mills\thanks{Now at LSU. Email: keith.mills@lsu.edu}\\
University of Alberta\\
{\tt\small kgmills@ualberta.ca}
}
\maketitle

\begin{abstract}
    \input{src/abstract}

\end{abstract}

\input{src/intro}

\input{src/method}

\input{src/datasets}
\input{src/results}

\input{src/related}

\input{src/conclusion}

\clearpage

{\small
\bibliographystyle{ieee_fullname}
\bibliography{egbib}
}

\newpage

\end{document}

%% file: src/abstract.tex
Pre-training is a general method that is used in a %
range of deep learning tasks. %
By first training a model on %
one task, and then further training on the downstream task used for final evaluation, the model is forced to learn a more general understanding of the input data. %
While pre-training has been applied to 3D Human Pose Estimation (HPE) previously, the scope of datasets used is typically very limited to some strong benchmarks, like Human3.6M. Therefore, in this project, we expand the scope of an existing 3D HPE scheme to be compatible with additional 2D and 3D HPE datasets, like Occlusion Person. We perform an extensive study on how aspects of 2D pre-training, such as model size, affect downstream performance, and to what extent pre-training can help the model generalize to different datasets. Experimental results show that 2D pre-training consistently outperforms training on 3D data alone, particularly in terms of computational efficiency. Finally, using MPII and Human3.6M, we are able to obtain %
an MPJPE score of under 64.5mm. We open-source our code fork here: \url{https://github.com/ECE740F21T01/pytorch-pose-hg-3d}

%% file: src/intro.tex
\section{Introduction}
\label{sec:intro}

Human Pose Estimation (HPE)~\cite{zheng2020deep} is concerned with the estimation of joint locations, e.g., a person's leg, arm or neck, from input data, primarily in the form of an image. %
2D HPE~\cite{zhang2019fast, luvizon20182d, fan2015combining, sun2017compositional} only considers the X and Y coordinates of a joint, while 3D HPE~\cite{Zhou_2017_ICCV, kocabas2019epipolar, kocabas2019vibe, sharma2019monocular, zhao2019semantic} %
predicts the depth, %
information that is not captured by an everyday camera and is therefore more difficult to infer.\footnote{This work was completed as a graduate course project more than four years prior to this preprint. It is shared for archival and educational purposes, primarily as an example of the level of technical writing, organization, and experimental reporting expected of senior undergraduate and early graduate coursework. The presentation reflects coursework-oriented expectations for clarity and completeness rather than the concision typical of peer-reviewed CS manuscripts. Portions of the experimental code were adapted and extended in subsequent research by the authors~\cite{mills2023aiop, salameh2024autogo}.}

\begin{figure}[t]
  \centering
  \subfloat[][Forearm clips into upper arm]{\includegraphics[width=1.5in]{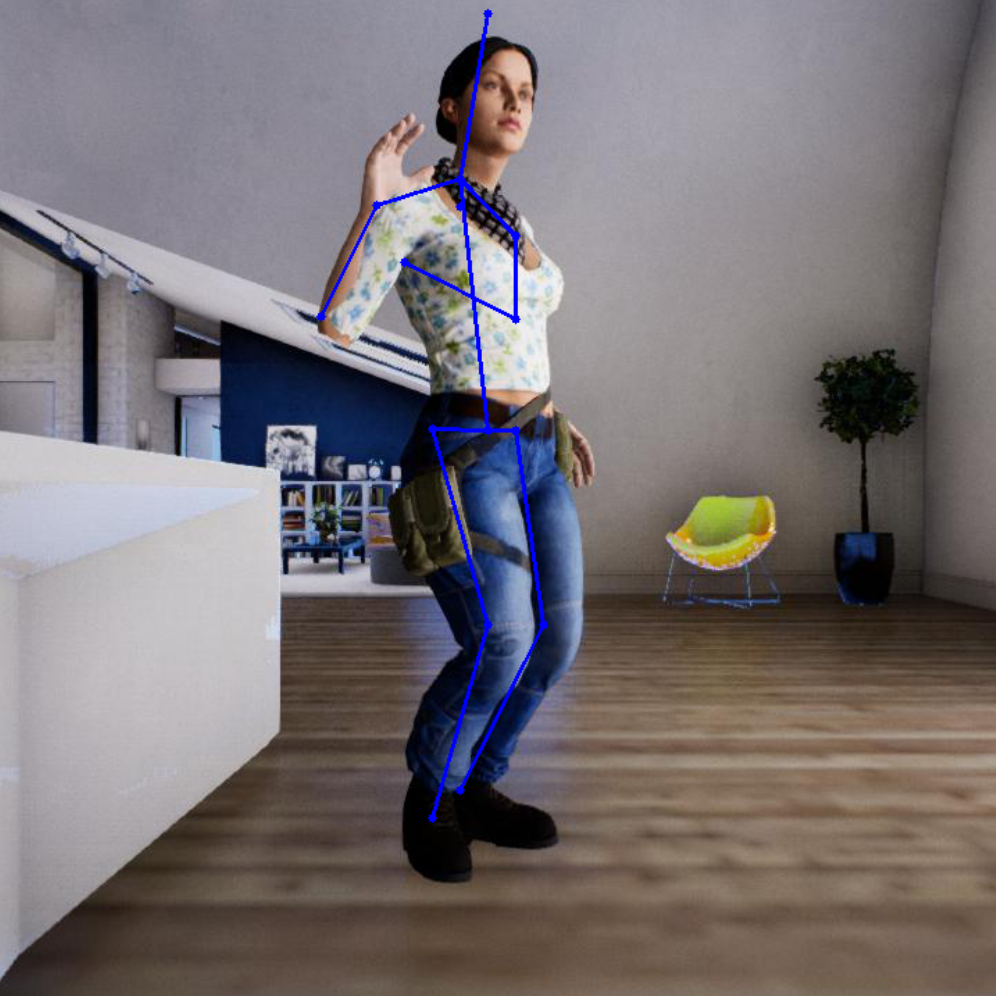}}\hspace{0.5mm}%
  \subfloat[][Arms clip into vest]{\includegraphics[width=1.5in]{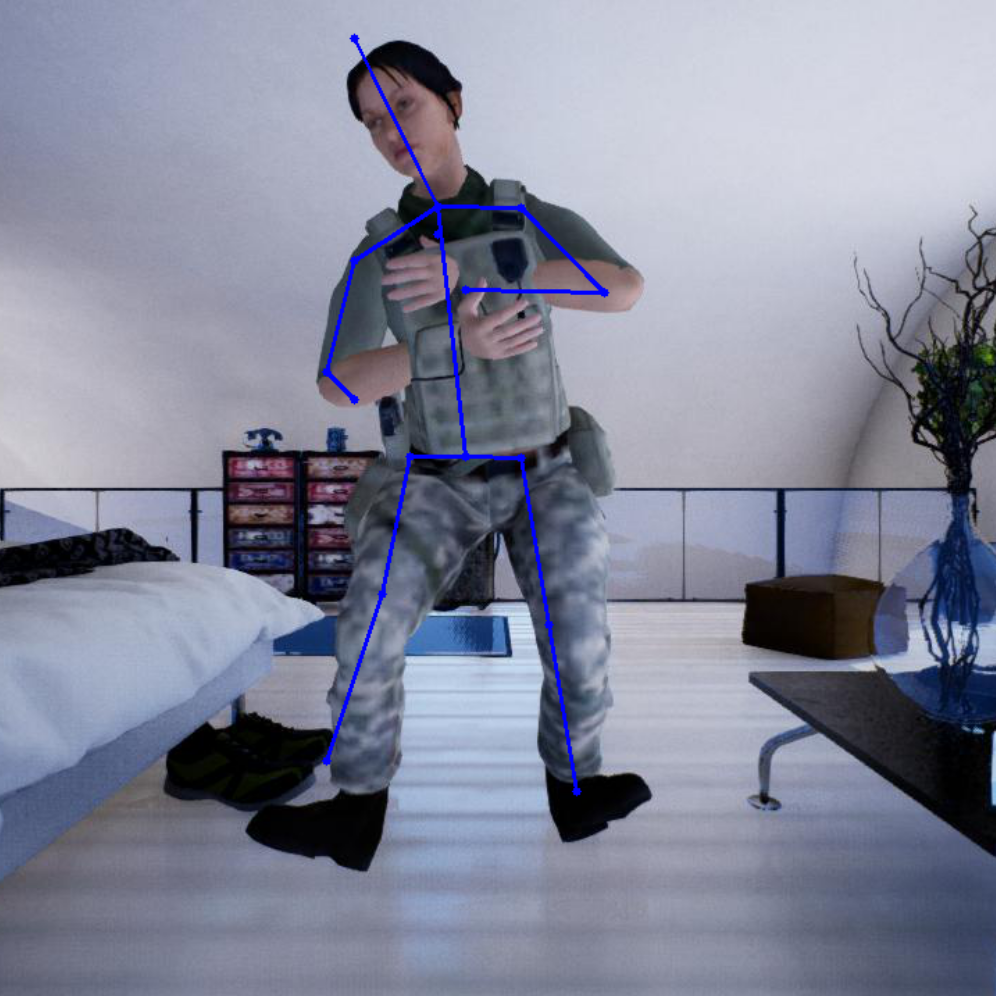}}%
  \caption{Example of the trained, publicly available model by \cite{Zhou_2017_ICCV} inferring the 3D joint locations of two sample images from Occlusion Person~\cite{occlusion_person}. Note how the actor's right forearm clips into another part of the body/attire, confusing the HPE model as to the actual length of the arm, and by extension, joint location.}
  \label{fig:occlusion_example}
  \vspace{-5mm}
\end{figure}

In general, one can grasp a sense of the additional difficulty associated with 3D HPE by understanding how HPE data is collected and annotated with joint labels. There are many 2D datasets in existence, including FLIC-Full~\cite{flic_full}, where movie screenshots are captured and annotations are then generated using Amazon Mechanical Turks (AMT). By contrast, the depth dimension of 3D HPE data necessitates the use of external Motion Capture (MoCap) devices %
to generate labels. MoCap equipment is generally worn by the actor making the pose and as such 3D HPE data is collected in controlled environments, as is the case in Human3.6M~\cite{h36m_pami, human36m_iccv}, a large 3D dataset and common benchmark. %

This requirement imposes several %
conditions on the quality of 3D HPE data, 
because poses are, in the strongest sense of the term, generated by `actors'; people performing a role. 
This isn't the case for 2D HPE as images of people in arbitrary, undirected poses can be labelled by human hand. As such, some 2D HPE data has come to be known by a specific term, `In the Wild', meant to encapsulate how the stochastic noise associated with the real world is incorporated with 2D datasets like Leeds Sports Pose-Extended (LSP), which uses images from Flickr~\cite{lsp_extended}, but is not captured to the same extent in 3D HPE data. 

In response, some 3D HPE methods~\cite{kocabas2019epipolar, kocabas2019vibe, sharma2019monocular} incorporate pre-training in order to improve model performance. Pre-training involves first fitting a model to a source task, and then transferring and fine-tuning it on a target task. Typically, the pre-training and downsteam tasks are not exactly the same, or the data used for each is different, forcing the model to learn a broader, more general representation of the input data. %

In particular, Zhou et al., 2017~\cite{Zhou_2017_ICCV} take 2D pre-training a step further. First, a ResNet-based encoder~\cite{he2016deep} learns to predict 2D HPE information, before the model learns on a fused dataset consisting of both 2D and 3D images, predicting all X and Y annotations using a shared module while separately estimating 3D depth annotations. %
A final stage of fine-tuning on the fusion dataset is  %
performed, involving the generation of estimated depth coordinates for the 2D samples from existing ground truth labels. The artificial 2D depth information is then learned through a novel geometric loss. However, such methods usually limit the datasets taken into consideration, focusing on one of the larger 2D datasets, MPII and the benchmark 3D dataset, Human3.6M.

Additionally, the limitations of 3D HPE datasets has motivated the creation of new, synthetic 3D HPE datasets like Occlusion Person~\cite{occlusion_person}. %
Utilizing the computational engines of video games, e.g. UnrealCV, %
virtual human figures and pose measurements can be generated in large quantities. %
However, these datasets suffer from a few limitations, such as improper representation of %
physical object collision, as Figure~\ref{fig:occlusion_example} illustrates.

Therefore, in this paper, we perform a study of the effectiveness of 2D pre-training for 3D HPE when multiple datasets are considered, as well as comparison of the effectiveness of 2D and synthetic 3D HPE datasets. Our contributions, specifically, are as follows: 

First, we extend the scope of Zhou et al., 2017~\cite{Zhou_2017_ICCV} to be compatible with the 2D datasets of FLIC-Full~\cite{flic_full} and LSP-Extended~\cite{lsp_extended} as well as the 3D dataset Occlusion Person~\cite{occlusion_person} and provide full integration of MPI-INF-3DHP~\cite{muco3dhp}. Originally, this method focuses on MPII~\cite{mpii} and Human3.6M~\cite{h36m_pami, human36m_iccv} as the primary 2D and 3D datasets, respectively. Integrating new HPE datasets into different methods is trickier than other image-based tasks like image classification on CIFAR-10~\cite{Krizhevsky09CIFAR} or ImageNet~\cite{deng2009imagenet} as the number of joints used, as well as joint selection within the human body, varies across different datasets. %
Nevertheless, by increasing the number of datasets compatible with our chosen method, we are able to provide a more thorough study of the effect of pre-training as well as comment on the effect of dataset size, as all three of our 2D datasets %
contain a different number of sample images. We show that using all three of our 2D HPE datasets, we are able to achieve MPJPE scores below 100 on Human3.6M, the minimum being under 64.5. %

Second, we perform a number of experiments comparing different selections of 2D and 3D data for model training and evaluation, as well as measure the performance of a model trained on 3D HPE data exclusively. By measuring execution time as well as performance, we are able to not only quantify the performance gains of 2D pre-training, but savings in terms of resource cost. Our experimental findings demonstrate that 2D pre-training is more resource efficient than training on 3D data only, which can takes hours longer to execute and be less generalizable to other datasets.

Third, we adjust the epochs in our training scheme based on sample discrepencies within each 2D dataset, in an attempt to improve the performance of smaller 2D datasets like FLIC-Full so that they may be more competitive with larger sets, e.g., MPII. We show that while performance improvements can be obtained, epochs are no substitute for the additional generalizability provided by a larger number of sample images.

The rest of this paper is organized as follows: In Section~\ref{sec:method} we describe the method of Zhou et al., 2017~\cite{Zhou_2017_ICCV} before enumerating our chosen datasets and listing what changes were needed to make them compatible with the method in Section~\ref{sec:data}. Then, we report a plethora of experimental findings and perform analysis on them in Section~\ref{sec:results} before discussing some related work and alternative methods %
in Section~\ref{sec:related}. Finally, we conclude %
in Section~\ref{sec:conclusion}.

%% file: src/method.tex
\section{Method}
\label{sec:method}

We adopt the method of 
Zhou et al., 2017~\cite{Zhou_2017_ICCV} to perform our experiments. In this section we summarize their procedure. 
The overall architecture consists of two modules. The first is a stacked hourglass network~\cite{newell2016stacked} built using a ResNet~\cite{he2016deep} backbone\footnote{By default, ResNet50 is used.} that performs 2D pose estimation. %
The second module takes the latent data representation of the hourglass network and performs regression on the the joint depth.

\subsection{2D Pose Estimation}
For a dataset with \textit{J} joints, the hourglass network receives 2D images as input and produces %
\textit{J} low-resolution heat-maps. %
Each heat-map represents the probability distribution for the location of a %
joint. The loss is not generated on the 2D labels directly. Rather, for a given joint %
annotation $Y$, %
we pass the label through a Gaussian filter $G$ %
to relax it from a single set of coordinates into a probability heat-map. The loss on the predicted heat-map $\hat{Y}$ is computed as,

\begin{equation}
    \centering
    \label{eq:2d_loss}
      \mathcal{L}_{2D}=\sum_{h}^{H}\sum_{w}^{W}(\hat{Y}^{(h,w)}-G(Y)^{(h,w)})^2,
\end{equation}
where $H$ and $W$ correspond to the height and width of the probability heat-maps, respectively.

\subsection{3D Depth Regression}
Two novel methods are used to learn 3D joint locations. The first is to train a network using 2D and 3D HPE data \textit{concurrently}. Second, %
a geometric loss which generates depth labels from 2D HPE labels.

\textbf{Pure 3D Pose Prediction.} For an arbitrary 3D HPE image, the loss on the 2D coordinates of each annotation is generated using Equation~\ref{eq:2d_loss}. %
For the predicted depth dimension $\hat{Y}_{dep}$ and ground-truth $Y_{dep}$, we compute the \textit{Smooth L1 Loss\footnote{Although \cite{Zhou_2017_ICCV} states use of a squared loss in the original paper, the code implementation at the time of our fork uses the Smooth L1 Loss.}}~\cite{girshick2015fast}, defined as %

\begin{equation}
    \centering
    \label{eq:3d_loss}
    \mathcal{L}_{dep, 3D} = 
    \begin{cases}
        0.5(\hat{Y}_{dep} - Y_{dep})^2 / \beta & \text{if $\abs{\hat{Y}_{dep} - Y_{dep}} < \beta$,} \\
        \abs{\hat{Y}_{dep} - Y_{dep}} - 0.5*\beta & \text{otherwise},
    \end{cases}
\end{equation}
where $\beta$ is typically set to 1.0. In contrast to a more traditional Mean Squared Error (MSE) loss, the Smooth L1 Loss is designed to control exploding gradients.

\textbf{3D Geometric Constraint Induced Loss.} The length of bones in the human skeleton have relatively fixed ratios to one another~\cite{Zhou_2017_ICCV, zheng2020deep}. Based on this assumption, it is possible to compute a depth-based loss on 2D HPE data. %
This loss is weakly-supervised, and acts to regularize the depth prediction module. 

Specifically, for a given set of bones $R_i$, where, for example, $R_{arm}$ consists of the major bones in the upper arms and forearms, we let $l_e$ correspond to the predicted length of bone $e \in R_i$, generated from the joint predictions. Next, we let $\bar{l}_e$ correspond to the bone length in a canonical skeleton, e.g. \cite{Zhou_2017_ICCV} use the average of all training subjects in Human3.6M~\cite{h36m_pami, human36m_iccv}. Using this \textit{a priori} knowledge on bone lengths, the following %
geometric loss is computed,

\begin{equation}
    \centering
    \label{eq:geometric_loss}
    \mathcal{L}_{geo} = \sum_{i}\frac{1}{\abs{R_i}}\sum_{e \in R_i}(\frac{l_e}{\bar{l}_e} - \frac{1}{\abs{R_i}}\sum_{e' \in R_i}\frac{l_{e'}}{\bar{l}_{e'}})^2.  
\end{equation}

Essentially, for each bone $e$ in a bone group $R_i$, we minimize the difference in bone length ratios compared to all other bones in the group, iterating over all $i$ groups. Due to the nature of Equations~\ref{eq:2d_loss} and \ref{eq:geometric_loss} especially, it is important all datasets share the same number of joints, and that those joints roughly correspond to the same locations in the body.

\subsection{Overall Loss}

In total, we use three different loss functions to train a model: One for 2D HPE, one for 3D depth regression, and one for 2D geometric depth regression. The overall loss on 3D depth is defined as,

\begin{equation}
    \centering
    \label{eq:overall_3d}
    \mathcal{L}_{dep}=
    \begin{cases}
    \lambda_{reg}\mathcal{L}_{dep,3D}, & \text{if $I \in \mathcal{I}_{3D}$,} \\
    \lambda_{geo}\mathcal{L}_{geo}, & \text{if $I \in \mathcal{I}_{2D}$,} \\
    \end{cases}
\end{equation}
where $I$ is a given image, while $\lambda_{reg}$ and $\lambda_{geo}$ are hyperparameters weighing both losses. $I \in \mathcal{I}_{3D}$ and $I \in \mathcal{I}_{3D}$ denote membership to 3D and 2D HPE datasets, respectively. Finally, combining the losses in above equations, the overall loss function for the entire network during fusion training on 2D and 3D HPE data concurrently is 

\begin{equation}
    \centering
    \label{eq:overall}
    \mathcal{L}=\mathcal{L}_{2D} + \mathcal{L}_{dep}.
\end{equation}

In %
practice, network training is not performed all at once. %
Zhou et al.~\cite{Zhou_2017_ICCV} conclude that %
the 2D and depth prediction modules have a dependency between them, %
that %
the geometric loss is highly non-linear, and %
consequently, use a more stable and effective three stage training procedure.  %
\textit{Stage 1} performs initialization of the 2D Pose Estimation module only with 2D annotated images using Equation~\ref{eq:2d_loss}. %
\textit{Stage 2} %
initializes the %
Depth Regression module using both 2D and 3D data, using Equation~\ref{eq:overall_3d} with $\lambda_{geo} = 0$. %
We further modify this stage to perform 3D exclusive training using Equation~\ref{eq:3d_loss} if desired. %
Finally, \textit{Stage 3} consists of fine-tuning %
the %
network with all data with geometric constraint; $\lambda_{geo} > 0$. Likewise, in our case, %
a 3D exclusive training option is integrated.

%% file: src/datasets.tex
\section{Datasets}
\label{sec:data}

This section summarizes the 2D and 3D HPE datasets we consider. We also enumerate the challenges involved in working with different HPE datasets, such as how to infer coordinates of missing joints and the scaling issue of 3D joint depth values, as that can lead to model divergence and poor performance if not scaled correctly.

\subsection{2D HPE Datasets}

Most 2D HPE datasets have their own body joints definition and have different number of joints in the annotation. To implement an approach that works universally across %
different datasets, we follow \cite{Zhou_2017_ICCV} and convert the annotations of new datasets. %

\textbf{MPII~\cite{mpii}} is the largest 2D HPE dataset we consider. %
It contains around 25k ``in the wild" images in total, divided into 22k and 3k subsets for training and testing, respectively. %
The images are extracted from YouTube videos and %
annotated by humans. %
There are $J = 16$ joints in total. Finally, MPII is the default 2D dataset %
\cite{Zhou_2017_ICCV} use to train the 2D pose estimation module. %
    
\textbf{Leeds  Sports  Pose-Extended  (LSP-Extended, LSP)~\cite{lsp_extended}} %
contains sports images collected from Flickr. %
Each image comes with annotation labels gathered using AMT. Following~\cite{lsp_extended}, we combine the 10k training images of LSP-Extended with the 1k training images and 1k testing images of the original Leeds Sports Pose %
dataset~\cite{lsp}, resulting in 12k total images, split training and test sets with 11k and 1k images, respectively. %

We borrow our pre-processing technique for LSP from%
~\cite{unipose}, %
which involves converting the 14 joint annotations to the 16 joint format as follows: We infer the missing thorax joints %
by interpolating the left and right shoulder joints which are closest to the thorax. Next, we %
infer %
missing hip joint by interpolating the left and right hip joints. The annotations are not expected to be extremely accurate due to the AMT human annotation and the inference of missing joints position when doing the joint format conversion.
    
\textbf{FLIC-Full~\cite{flic_full}} is the smallest 2D HPE dataset. It only contains 5k %
total images split into training and test sets with 3k %
and 1k %
images, respectively. %
The images used in this dataset come from movies 
which cause some of the joints to be occluded or invisible within the %
frame. The annotations are collected using AMT. %
Many occluded or invisible joints are annotated with not a number (NaN) values. %
We substitute %
any NaN values in the annotation %
for 0.

FLIC-Full %
uses a human skeleton format %
with 29 joints. It does not have annotations for 3 of the 16 joints (thorax, neck, headtop) %
used by MPII. %
Therefore, we infer the missing joint thorax, neck, and headtop joints annotations by using the coordinate of the FLIC-Full dataset middle torso, middle shoulder, and nose joints annotations respectively, as %
they are the closest to these missing joints physically.

\subsection{3D HPE Datasets}

Similar to our pre-processing for 2D datasets following~\cite{Zhou_2017_ICCV}, we convert the different body joints format of each 3D HPE datasets to the same MPII 16-joint format by inferring missing joints from neighbors %
, and we align the 3D joint values to the root (hip) joint by subtracting it.

For all 3D %
datasets, we require three types of annotations following \cite{Zhou_2017_ICCV}. First, we need the root-aligned 3D pose in the camera coordinate system for evaluation purpose. Second, the 2D pose in image coordinate system is necessary to train the 2D pose estimation module. Third, the scaled, root-aligned depth value relative to the image plane is needed to train the depth regression module. The first two values %
are given in %
each dataset.

For the root-aligned depth value, %
we need to find the depth scaling factor between the camera and image coordinate system. We calculate this depth scaling factor by solving a similarity transform between the root-aligned 3D pose and the 2D pose~\cite{sun2017integral,sun2018integral}. %
Correct depth-scaling of the annotations are critical for the depth regression module %
as incorrectly scaled depth annotations can lead to very high MPJPE values and model divergence.

\textbf{Human3.6M (H36M)~\cite{human36m_iccv, h36m_pami}} is a large-scale 3D HPE dataset %
which contains up to 3.6 million sample images. %
Dataset samples are taken indoors in a controlled environment and joint annotations are captured using a MoCap system. H36M is a widely used benchmark used throughout the literature~\cite{zheng2020deep, kocabas2019epipolar, kocabas2019vibe, sharma2019monocular}. %

Following~\cite{Zhou_2017_ICCV}, we use the ECCV 2018 Challenge subset of H36M which contains %
36k training images and %
19k test images, which is around 55k in total, %
and their corresponding 3D annotations. We adopt %
their processing scheme %
to convert %
the 17 body joints of H36M to the 16 joints of MPII~\cite{mpii}. The MPII thorax joint is missing in H36M and is replaced by the spine joint in H36M.

\textbf{MPI-INF-3DHP (MPII3D)~\cite{mono-3dhp2017}} contains %
967k training images and %
23k test images for 990k in total.  %
Poses are gathered from both indoor and outdoor environments. The dataset provides RGB images and their corresponding 3D body joints coordinates captured with a markerless MoCap system. %
There are 28 joints for the training set, and a 17 joints %
for the testing set, but neither contain a thorax joint. %
Therefore, like H36M, we infer the thorax joint by using the coordinates of the nearest spine joint when converting to MPII 16 joints format.

Although \cite{Zhou_2017_ICCV} report some results using the MPI-INF-3DHP test set, %
we did not find %
in their released code repository. %
Therefore, to use the training and test sets, %
we adopt the %
pre-processing procedure of %
VIBE~\cite{kocabas2019vibe}.

\textbf{Occlusion Person (OP)~\cite{occlusion_person}} is a synthetic 3D human pose dataset which contains %
264k training images and %
130k test images, for around 394k %
in total. OP %
uses the open-source UnrealCV game engine to create RGB images and depth maps of 3D human models. OP %
puts objects such as sofas and desks to occlude part of the human to create poses with %
occluded joints. It provides 3D ground-truth coordinates %
for 15 body joints, %
not including the thorax and headtop joints of MPII. %
To convert it to the MPII 16 joint format, we infer the coordinate for the thorax joints by interpolating the left and right shoulder joints, and we use the neck joint coordinates for the missing headtop joint.

\subsection{Metrics}

We adopt two simple, %
common %
metrics to quantify performance. For 2D HPE, we use the Percentage of Correct Points (PCK) measurement, which is an accuracy measure computed by comparing joint predictions to the ground-truth within a threshold. Specifically, we use PCKh@0.5, where the threshold for determining a correct joint is 50\% of the head segment length. This measure is reported as a percentage and \textit{higher} is better.

For 3D HPE, we use Mean Per Joint Position Error (MPJPE), ``the most widely used evaluation metric"~\cite{zheng2020deep}.  %
We compute it by taking the Euclidean distance between an estimated 3D joint and the ground-truth location, and averaging the result over all joints in a pose. This metric is reported in millimeters [mm] and \textit{lower} is better.

%% file: src/results.tex
\begin{table*}[t]%
    \centering
    \scalebox{0.95}{
    \begin{tabular}{cc|c|cc|cc|ccc} \toprule
        \multicolumn{2}{c|}{\textbf{Training Set}} & \textbf{Stage 1} & \multicolumn{2}{c|}{\textbf{Stage 2}} & \multicolumn{2}{c|}{\textbf{Stage 3}} & \multicolumn{3}{c}{\textbf{Test MPJPE}} \\ \midrule
        \textbf{2D} & \textbf{3D } & \textbf{2D PCKh} & \textbf{2D PCKh} & \textbf{3D MPJPE} & \textbf{2D PCKh} & \textbf{3D MPJPE} & \textbf{H36M} & \textbf{MPII3D} & \textbf{OP} \\ \midrule
         MPII~\cite{Zhou_2017_ICCV} & -- & -- & -- & -- & -- & -- & 64.55 & -- & -- \\
         MPII (Test) & H36M & -- & -- & -- & -- & -- & 64.55 & 179.05 & 200.96 \\ \midrule
         MPII  & H36M & 85.59\% & 94.45\% & 63.54 & 94.52\% & 63.93& \textbf{64.49} & 185.52 & 207.03 \\
         MPII  & MPII3D & -- & 9.59\% & 362.21 & 9.86\% & 366.76 & 131.39 & 175.89 & 205.65 \\
         MPII  & OP & -- & 21.04\% & 323.91 & 21.06\% & 323.98 & 158.05 & 207.12 & \textbf{118.91} \\
         LSP & H36M & 68.48\% & 92.32\% & 73.96 & 92.33\% & 73.25 & 73.63 & \textbf{171.69} & 234.23 \\
         LSP & MPII3D & -- & 8.98\% & 372.89 & 8.75\% & 368.44 & 177.48 & 183.87 & 273.52 \\
         LSP & OP & -- & 22.11\% & 318.27 & 22.14\% & 324.03 & 202.96 & 222.31 & 146.91 \\
         FLIC & H36M & 56.67\% & 85.91\% & 98.64 & 85.47\% & 99.60 & 96.81 & 418.77 & 503.01 \\
         FLIC & MPII3D & -- & 9.42\% & 376.68 & 8.89\% & 378.75 & 437.86 & 351.69 & 491.21 \\
         FLIC & OP & -- & 25.76\% & 298.91 & 24.88\% & 89.29 & 475.46 & 483.29 & 191.42 \\ \bottomrule
    \end{tabular}
    }
    \caption{Results for different combinations of 2D and 3D datasets. %
    The first row is the reported performance of~\cite{Zhou_2017_ICCV}. For the %
    second row of results, we evaluated the publicly available, pre-trained weights on all 3D datasets manually. We re-use Stage 1 models for later stages. For Stage 1, 2 and 3 results, we report metrics on the validation set, which is defined as 10\% of the test set.}
    \label{table:reproduction}
\end{table*}

\section{Experimental Results}
\label{sec:results}

In this section, we describe our experimental procedure and then enumerate our findings. First, we replicate the results of \cite{Zhou_2017_ICCV} and compare to the originally published metrics. Next, we perform several experiments using our selected 3D HPE datasets only, reporting their performance in the absence of 2D pre-training or fusion training. %
Finally, %
we attempt to improve accuracy and MPJPE performance by taking dataset size into account. %

\subsection{Reproduction of Results}
\label{sec:repro}

As a starting point, we first attempt to reproduce the original results of \cite{Zhou_2017_ICCV} using the directions prescribed in the code repository. %
The `out-of-the-box' hyperparameters for each stage are as follows:

\begin{enumerate}
    \item \textbf{Stage 1 (S1)}: Model trains %
    for 140 epochs with a batch size of 32. The Adam~\cite{kingma14Adam} optimizer is used with an initial %
    learning rate of 0.001, which is reduced by a factor of 10 at the 90th and 120th epochs. %
    \item \textbf{Stage 2 (S2)}: Fusion training is performed with %
    $\lambda_{reg} = 0.1$. %
    The model is trained for 60 epochs and the learning rate is reduced at epoch 45. All other hyperparameters are unchanged from S1.
    \item \textbf{Stage 3 (S3)}: %
    $\lambda_{geo}$ is set to 0.01. The model %
    trains for 10 epochs at a learning rate of 0.0001. All unmentioned hyperparameters remain unchanged from S3.
\end{enumerate}

It is important to note that when performing fusion training with both 2D and 3D datasets, by default, there is a 1:1 ratio of the number of 2D and 3D samples. That is, only a subset of the total 3D images are used. Likewise, evaluation on the validation set is only performed every 5 epochs. 

Table~\ref{table:reproduction} reports our initial results as a cartesian product of all 2D and 3D HPE datasets for training and 3D MPJPE evaluation. As can be seen, when we evaluate the provided, pre-trained weights on Human3.6M, we are able to match the originally reported performance. Moreover, we observe a slight, but nevertheless further improvement when retraining a model from scratch. 

The method, with its original hyperparameters, is able to generalize to other 3D HPE datasets somewhat, as evidenced by how the best MPJPE results for MPI-INF-3DHP and Occlusion Person are below 200mm. In fact, the viability of one of the 2D datasets we integrated with \cite{Zhou_2017_ICCV}, LSP-Extended~\cite{lsp_extended} is best shown when evaluating on MPI-INF-3DHP, as it achieves the best overall MPJPE. 

However, this phenomena does not show that LSP-Extended is \textit{always} better than MPII, rather that it \textit{can} be. In every other situation MPII outperforms LSP-Extended. %
Likewise, %
FLIC-Full~\cite{flic_full} always obtains the worst performance. %
Overall, the results observed are arguably the expected outcome: MPII is the largest dataset, so it obtains the best performance most of the time, LSP-Extended has a medium number of entries and is ranks 2nd on most evaluations, while FLIC-Full, due to its small size, obtains the worst results, though it still obtains an MPJPE score of under 100 on Human3.6M. We again note the fixed number of epochs used in these experiments, %
and attempt to alleviate the disadvantages associated with having a small number of 2D samples in Section~\ref{sec:improve}. %

\begin{table*}[t]
    \centering
    \scalebox{0.95}{
    \begin{tabular}{cc|cc|cc|ccc} \toprule
        \multicolumn{2}{c|}{\textbf{Training Set}} & \multicolumn{2}{c|}{\textbf{Stage 2}} & \multicolumn{2}{c|}{\textbf{Stage 3}} & \multicolumn{3}{c}{\textbf{Test Evaluation MPJPE}} \\ \midrule
        \textbf{Stage 2} & \textbf{Stage 3} & \textbf{2D PCKh} & \textbf{3D MPJPE} & \textbf{2D PCKh} & \textbf{3D MPJPE} & \textbf{H36M} & \textbf{MPII3D} & \textbf{OP} \\ \midrule
        H36M & H36M & 72.01\% & 138.63 & 71.59\% & 139.48 & 137.58 & 535.40 & 510.21 \\
        H36M & MPII3D & -- & -- & 7.52\% & 445.30 & 444.07 & 319.24 & 501.77 \\
        H36M & OP & -- & -- & 24.98\% & 329.99 & 332.24 & 385.08 & 132.20 \\
        MPII3D & MPII3D & 7.76\% & 368.33 & 8.50\% & 338.28 & 342.29 & 249.69 & 382.74 \\
        MPII3D & H36M & -- & -- & 88.47\% & 86.58 & \textbf{86.43} & 456.84 & 481.32 \\
        MPII3D & OP & -- & -- & 24.96\% & 297.29 & 297.70 & 329.28 & \textbf{111.55} \\
        OP & OP & 27.86\% & 295.38 & 29.26\% & 311.23 & 312.71 & 338.24 & 150.43 \\
        OP & H36M & -- & -- & 84.11\% & 103.50 & 104.00 & 427.27 & 270.80 \\
        OP & MPII3D & -- & -- & 9.45\% & 297.13 & 299.14 & \textbf{198.69} & 402.79 \\ \bottomrule
    \end{tabular}
    }
    \caption{Found results when 3D HPE datasets are used exclusively to train the model. For Stages 2 and 3, we perform tests on all possible combinations of our selected 3D datasets. %
    We report the 2D PCKh@0.5 accuracy and 3D MPJPE results on the validation set.} 
    \label{table:exclusive_3d}
    \vspace{-3mm}
\end{table*}

\subsection{Using 3D Data Exclusively}
\label{sec:3dOnly}

For our second set of experiments, focus exclusively on on 3D training, starting from Stage 2. %
In this scenario, we omit Stage 1 of model training. For experiments on MPI-INF-3DHP and Occlusion Person, the size of the training set is much larger than the version of Human3.6M %
Zhou et al., 2017~\cite{Zhou_2017_ICCV} utilize. Therefore, for experiments with these datasets, we increase the batch size to 128 and limit the number of epochs. Specifically, for Stage 2, we run for 5 epochs and reduce the learning rate after epoch 3, and for Stage 3, we only run for 2 epochs. Also, in Stage 2 we set $\lambda_{reg} = 1.0$, then reduce it to 0.1 for Stage 3.

Results are listed in Table~\ref{table:exclusive_3d}. It should be noted that in this scenario, Equation~\ref{eq:geometric_loss} is not used due to a lack of 2D data. %
Rather, the focus is on fine-tuning the model with reduced learning rate and $\lambda_{reg}$, potentially on a different dataset than was used to train the model in Stage 2.

At first glance, we note that the best results obtained for each dataset \textit{do not} correspond to the instances where Stage 2 and 3 were both executed using the same dataset. Rather, the best results are achieved when the target data is used in Stage 3 for fine-tuning, but when one of the larger 3D datasets %
are used in Stage 2. This demonstrates the utility of the generalizability a model will learn when being trained on different datasets if not tasks. It also reinforces one of the findings we previously noted: That the size of the dataset used to pre-train the model has an impact on target performance. 

Nevertheless, we also note that overall, the MPJPE results in Table~\ref{table:exclusive_3d}, where only 3D data is used, are generally higher than those in Table~\ref{table:reproduction}, and therefore worse. On Human3.6M, the best result obtained using 3D data only is worse than the result obtained using 2D pre-training with MPII or LSP-Extended, and the best result for MPI-INF-3DHP is noticeably worse than the best 2D result. Only on Occlusion Person does the use of 3D data exclusively barely outperform the best 2D counterpart, and even then, this advantage comes at higher resource cost in terms of time and GPU memory.

Given the already specified hyperparameters, a single epoch of MPII will execute in 2.6 minutes in Stage 1, and when paired with Occlusion Person for Stages 2 and 3, the time per epoch increases to 5.0 and 7.6 minutes for Stages 2 and 3, respectively. Multiplying these times by the number of epochs in each stage, 140, 60 and 10, we get a total training time of approximately 740 minutes, or \textbf{12 hours and 20 minutes}\footnote{These times do not account for miscellaneous overhead such as the time to load a model, however, we argue this is negligible. The model size and structure does not change, so the time to load should be constant.}. By contrast, when running 3D training with MPII3D or Occlusion Person exclusively, each epoch of Stages 2 and 3 will take roughly 200 and 31.8 minutes to execute, respectively, for 1064 minutes or \textbf{17 hours, 44 minutes} of total training time. This is an increase of 5 hours over the best 2D pre-training result, for less than 10mm improvement in MPJPE. It is a very steep trade-off.

As another example, consider the best results on MPI-INF-3DHP. In Stage 1, it will take 1.13 minutes per epoch on LSP-Extended, then 2.45 and 3.25 minutes for Stages 2 and 3, respectively, for a total of 337.7 minutes of \textbf{5 hours, 38 minutes}. Again, we use MPI-INF-3DHP and Occlusion Person for the best 3D exclusive training model, only switching the order the datasets are used, for a total training time of 559 minutes or approximately \textbf{9 hours, 19 minutes}, an increase of almost 4 hours training time.

We again note that when training on either MPI-INF-3DHP or Occlusion Person exclusively, we do not train for very many epochs, and adding more would only further increase the time cost. The same can be said for using the same batch size as 2D pre-training and fusion training experiments; bringing it back down to 32 from 128 for 3D training exclusively would only increase the time needed to execute an experiment, albeit one would save GPU memory in doing so. Regardless, these findings demonstrate that the resource benefits of the 2D pre-training and fusion training scheme on the the 3D HPE datasets of MPII3D and Occlusion Person do not come at the cost of performance.

\begin{table}[t]
    \centering
    \scalebox{0.95}{
    \begin{tabular}{cc|ccc} \toprule
        \multicolumn{2}{c|}{\textbf{Training Set}} & \multicolumn{3}{c}{\textbf{Test MPJPE}} \\ \midrule
        \textbf{2D} & \textbf{3D} & \textbf{H36M} & \textbf{MPII3D} & \textbf{OP} \\ \midrule
        MPII & H36M & \textbf{64.49} & 185.52 & 207.03 \\ 
        MPII & H36M/MPII3D & 93.91 & 152.91 & 199.62 \\ 
        MPII & H36M/OP & 105.27 & 171.26 & \textbf{127.17} \\
        LSP  & H36M & 72.46 & 180.71 & 246.21 \\ 
        LSP  & H36M/MPII3D & 88.93 & \textbf{151.03} & 266.99 \\
        LSP  & H36M/OP & 140.14 & 189.71 & 143.69    \\ 
        FLIC & H36M & 88.58 & 417.87 & 452.63 \\
        FLIC & H36M/MPII3D & 115.98 & 232.15 & 412.09 \\
        FLIC & H36M/OP & 167.96 & 288.76 & 167.61 \\ \bottomrule
    \end{tabular}
    }
    \caption{2D pre-training results when the number of epochs used to run experiments for FLIC-Full and LSP-Extended is doubled. For 3D datasets, if two names are listed, the first dataset is used in Stage 2, and the second in Stage 3. MPJPE metrics in millimeters. The first row is directly copied from Table~\ref{table:reproduction} for ease of viewing.}
    \label{table:equalize}
\end{table}

\subsection{Improvement via Epoch Scaling}
\label{sec:improve}
In this section we perform an %
experiment to %
improve the performance observed when pre-training using FLIC-Full and LSP-Extended. Results in Section~\ref{sec:repro} show that when Human3.6M is selected to train the model, good performance can be observed on other 3D datasets, like MPI-INF-3DHP. Moreover, we can roughly rank the results obtained according to the size of the 2D dataset used; MPII is the largest and almost always obtained the best result, while FLIC-Full is the smallest and always got the worst result. 

Therefore, we perform additional experiments where Human3.6M is used exclusively in Stage 2 of model training, while all three 3D datasets can be selected to fine-tune the model in Stage 3. Additionally, when performing further experiments on FLIC-Full and LSP-Extended, we double the number of epochs in Stages 1, 2 and 3 to be 280, 120 and 20, respectively\footnote{Epochs where the learning rate is reduced are adjusted accordingly.}. MPII contains 22k training images while LSP-Extended, the 2D dataset most likely to be competitive with it, contains 11k training images, so doubling the epochs essentially equalizes the number of batches passed through the model and backpropagation updates applied.

In performing this experiment, we answer two additional questions. First, to what extent does increasing the number of epochs improve performance? Second, if performance is improved, is %
this method %
an effective substitute for using a larger, more diverse dataset in terms of poses and labels? Holding all unmentioned hyperparameters as-is, we compile our findings in Table~\ref{table:equalize}. 

First off, a trivial finding that performance on MPII improves for Human3.6M when that dataset is used for the majority of 3D training in Stage 2, and we even observe some minor decreases in MPJPE for MPI-INF-3DHP, but not in all scenarios. Second, doubling the epochs for LSP-Extended slightly improves Human3.6M performance when that dataset is used in Stages 2 and 3, but when the same model is evaluated on the other 3D datasets, MPJPE worsens. However, when other datasets are used in Stage 3, we observing vary degrees of improvement for all test scenarios. For FLIC-Full, all test MPJPE scores are improved compared to the findings of Table~\ref{table:reproduction}. Lastly, neither LSP-Extended or FLIC-Full manage to outperform MPII on Human3.6M.

\begin{figure}[t!]
  \centering
  \subfloat[][LSP Model Prediction]{\includegraphics[width=1.5in]{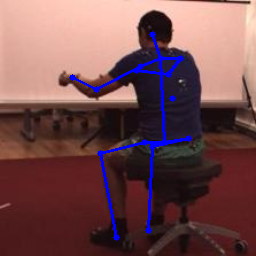}}\hspace{0.5mm}%
  \subfloat[][FLIC Model Prediction]{\includegraphics[width=1.5in]{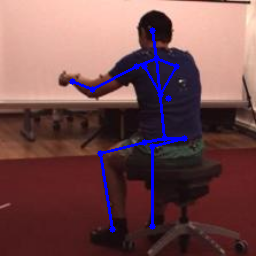}}\par
  \subfloat[][LSP Model Wireframe]{\includegraphics[width=1.5in]{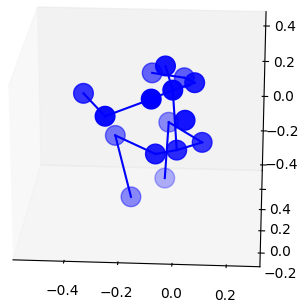}}%
  \subfloat[][FLIC Model Wireframe]{\includegraphics[width=1.5in]{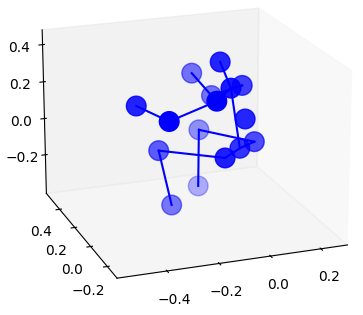}}\par
  \caption{Example of predicted joint annotations for a sample image from Human3.6M. Annotations were generated by LSP-Extended and FLIC-Full models trained for double epochs.}
  \label{fig:wireframes}
\end{figure}

Taken together, these findings indicate that while increasing the number of epochs can improve performance when using a smaller 2D dataset, that alone is no substitute for using a larger 2D dataset for pre-training. %
Additionally, using different 3D datasets during the training scheme tends to improve performance, and therefore, overall model generalizability in 3D HPE scenarios, even when the final test dataset is not used to train the model at all. 

Finally, taking Tables~\ref{table:reproduction}, \ref{table:exclusive_3d} and \ref{table:equalize} into account, we note that using Occlusion Person in training generally produces lackluster performance when generalizing to other 3D datasets. That is, if a model is trained on Human3.6M, it will usually obtain good performance on MPI-INF-3DHP, and vis-versa, but if a model is trained on Occlusion Person and possibly another 3D dataset, the corresponding MPJPE performance will be worse on unseen data. A clear example of this is shown using the Human3.6M evaluations in Table~\ref{table:equalize}, where models fine-tuned with MPI-INF-3DHP consistently outperform those fine-tuned on Occlusion Person, demonstrating a limit of synthetic 3D HPE data.

\begin{figure}
  \centering
  \subfloat[][]{\includegraphics[width=1.7in]{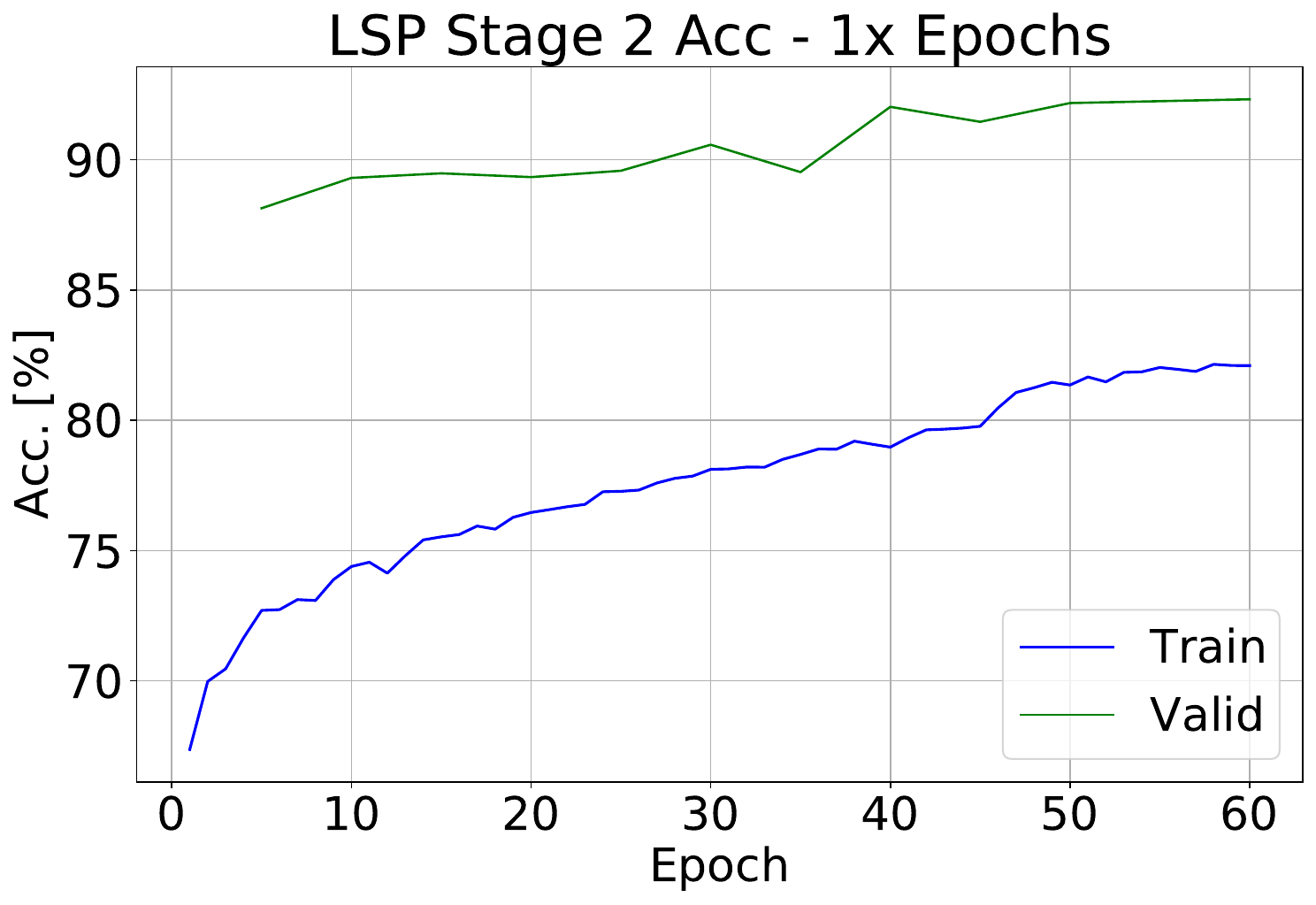}}%
  \subfloat[][]{\includegraphics[width=1.73in]{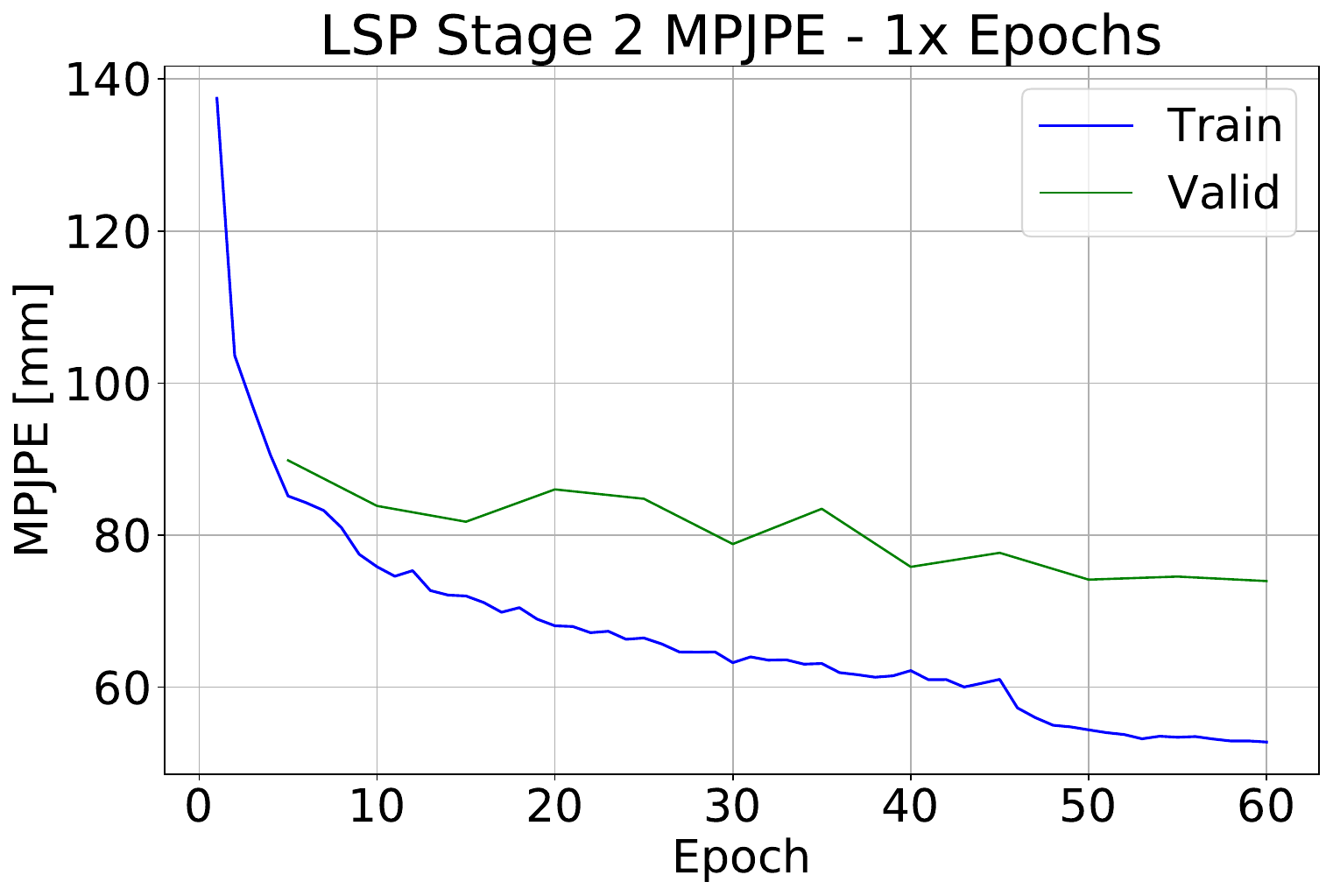}}\par
  \subfloat[][]{\includegraphics[width=1.7in]{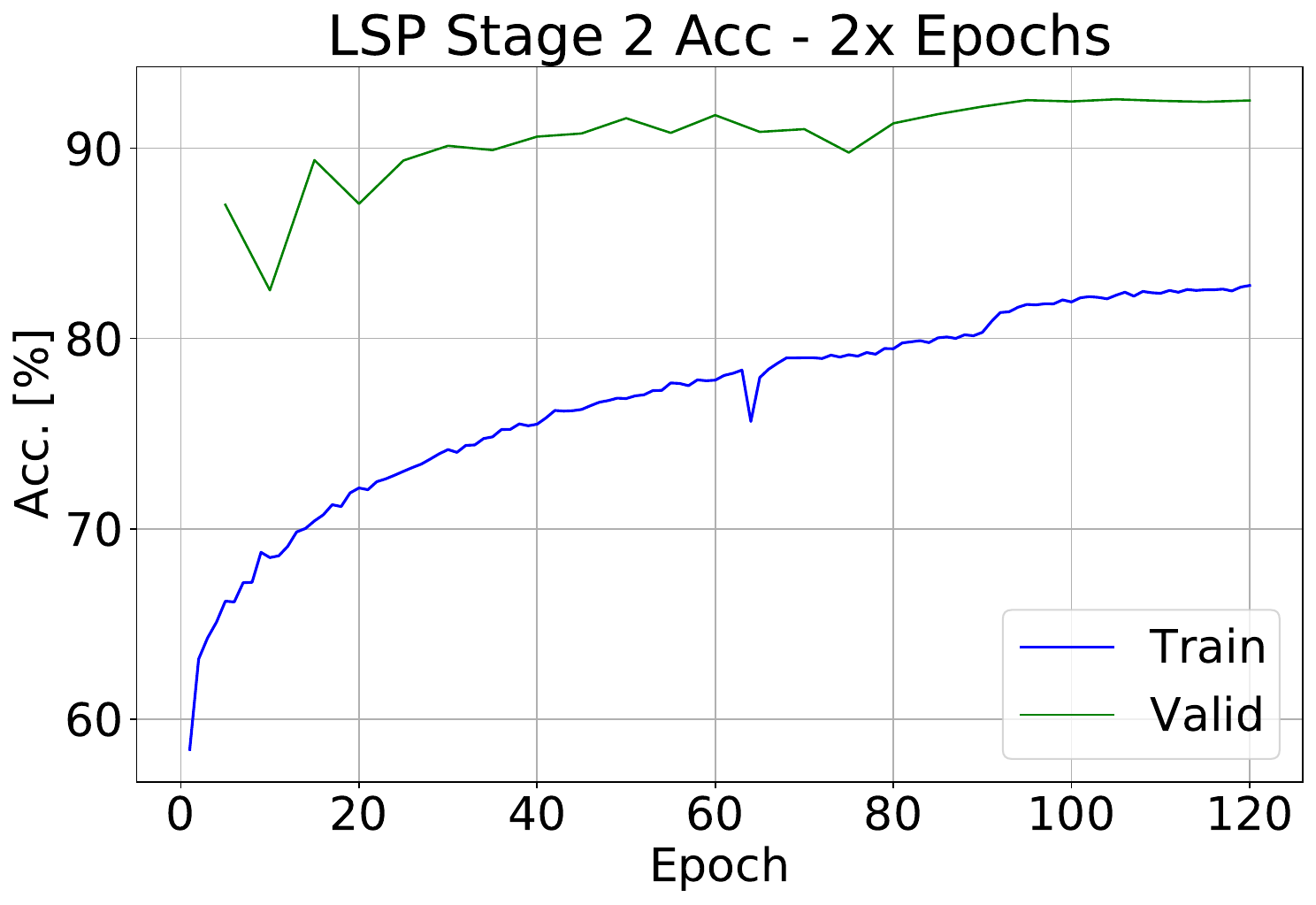}}%
  \subfloat[][]{\includegraphics[width=1.73in]{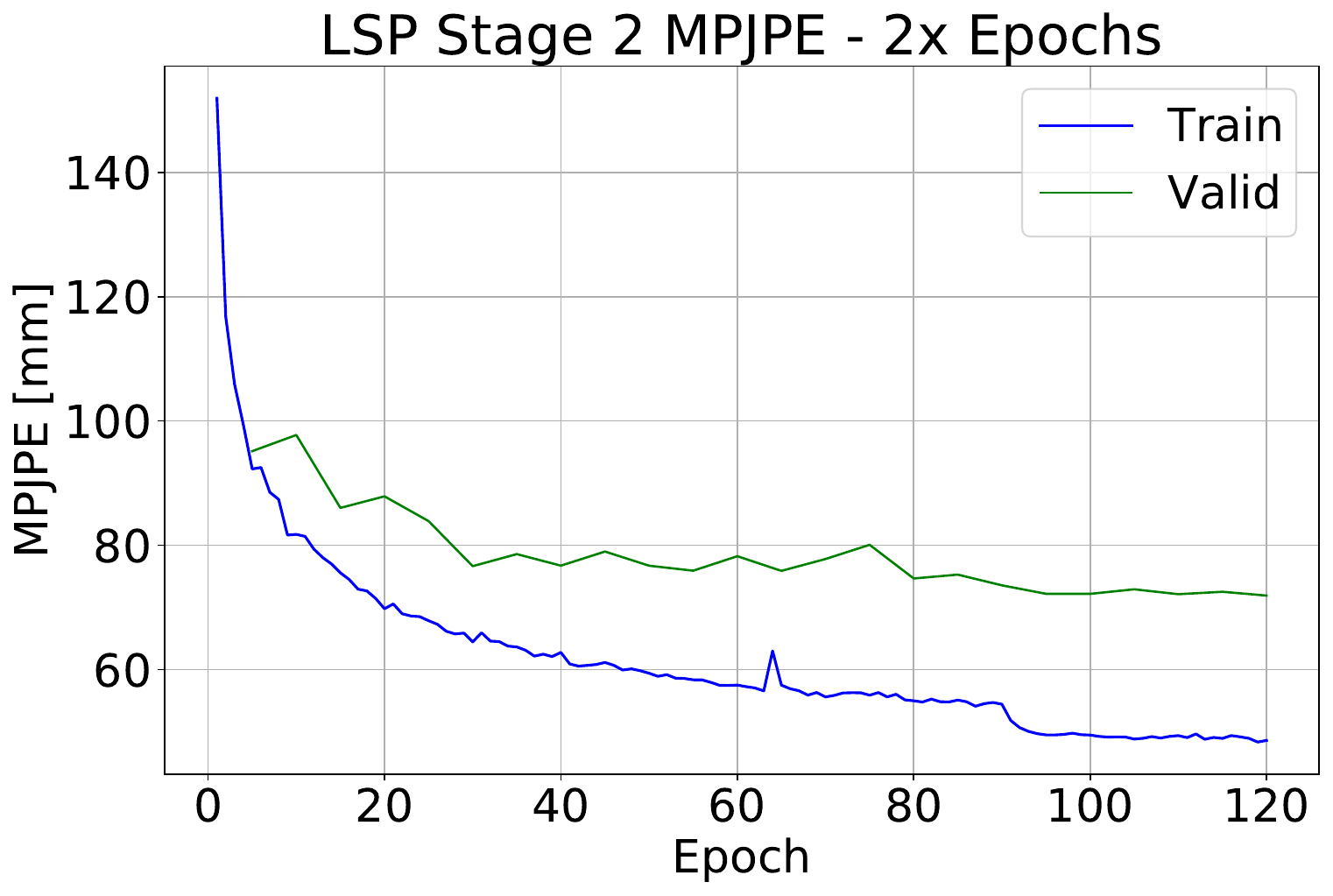}}\par
  \subfloat[][]{\includegraphics[width=1.7in]{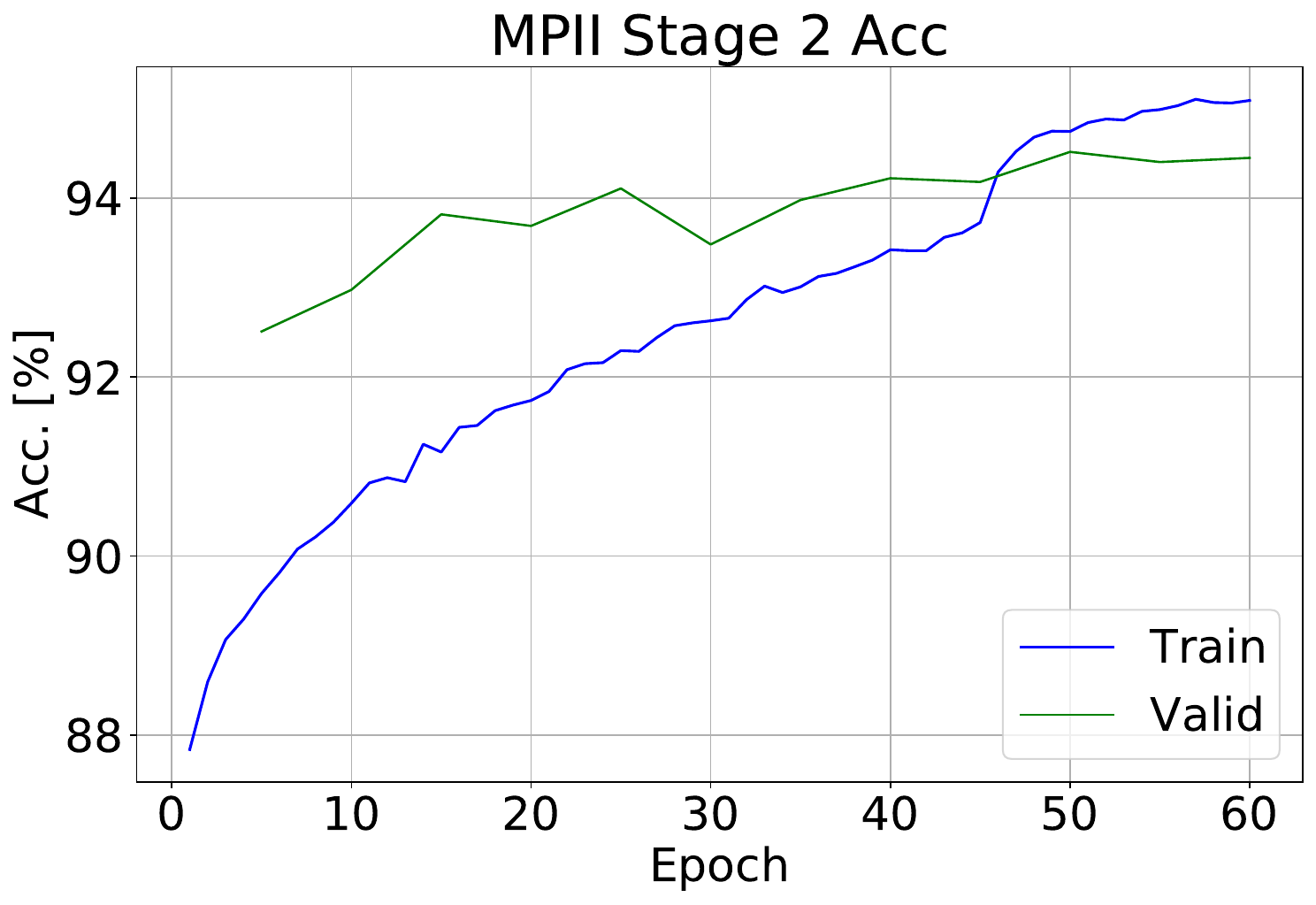}}%
  \subfloat[][]{\includegraphics[width=1.73in]{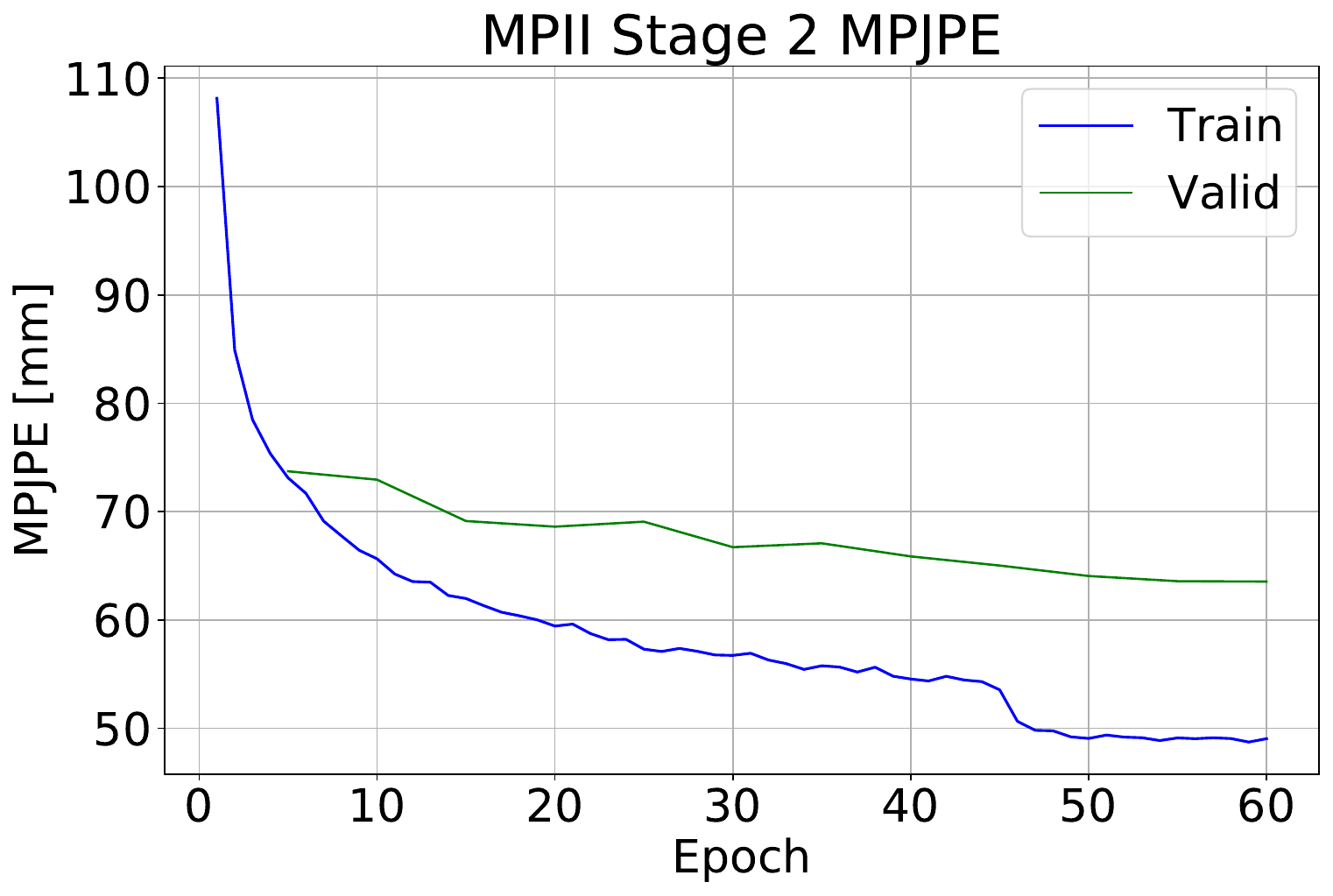}}%
  \caption{Stage 2 training plots comparing 2D PCKh@0.5 Accuracy (first column) and 3D MPJPE (second column) on the training and validation sets for three sets of experiments. The first row (Subfigures (a) and (b)) list LSP-Extended results from Section~\ref{sec:repro}. Subfigures (c) and (d) correspond to LSP-Extended when the number of epochs is doubled in Section~\ref{sec:improve} while (e)-(f) correspond to MPII results with the default hyperparameters.}
  \label{fig:3x2plots}
  \vspace{-4mm}
\end{figure}

Next, we provide example visualizations of the predicted joint locations our models trained on Human3.6M in Stage 2 and for twice the normal epochs can produce in Figure~\ref{fig:wireframes}. Note that in contrast to Figure~\ref{fig:occlusion_example}, one limb is fully occluded, rather than clipping into another part of the body or attire, and yet both models make different predictions for the true location of the hidden right arm. Moreover, these predictions are both reasonable, and upon visual inspection of the wireframe, preserve the ratio of arm and leg lengths. 

Additionally, Figure~\ref{fig:3x2plots} illustrates the accuracy and MPJPE curves of three different experiments in Stage 2: LSP-Extended and MPII with unchanged hyperparameters, and LSP-Extended with a scaled number of epochs. In Stage 2, the learning rate is reduced when three-fourths of epochs have completed, and we note that for all three experiments, there is a noticable increase in training accuracy and drop in training MPJPE. On the validation set these metrics do not rise or fall as dramatically, but they do settle closer to their final values, and no longer fluctuate as much. Finally, it is interesting to note the subtle changes in the LSP-Extended MPJPE curves: While both curves look to very similar to each other and ultimately finish at similar training and validation MPJPE values, %
when the learning rate drops, the regular curve has a training MPJPE of around 60mm, while the curve corresponding to the experiment with epoch scaling is noticeably lower, around 55mm.

%% file: src/related.tex
\section{Related Work and Alternatives}
\label{sec:related}

Pre-training is not exclusive to HPE tasks. For example, \cite{gnn_pretrain} pre-train Graph Neural Networks (GNN) for classification by predicting if a target node falls within a specific neighborhood, or masking the attributes of nodes and teaching the GNN to predict the values. Similar to our own work, \cite{zhang2018unreasonable} use the latent representation of input images from pre-trained network architectures, e.g. SqueezeNet~\cite{iandola2016squeezenet} or VGG~\cite{simonyan2014very} to learn the perceptual similarity of images. Closer to HPE, EpipolarPose~\cite{kocabas2019epipolar} and  \cite{sharma2019monocular} use a pre-trained CNNs that are not updated while learning HPE. One issue with using \cite{kocabas2019epipolar} is a reliance on custom-generated pose annotation files for which the authors did not provide code for. By contrast, we extended the framework of \cite{Zhou_2017_ICCV} by estimating or substituting unknown joints and update the 2D CNN during all stages of model training. 

Our list of 3D HPE datasets is not exhaustive. %
For example, we integrate 3DPW~\cite{3dpw} into our scheme, but ultimately do not report many results on it, because the annotations are missing too many joints compared to the baseline Human3.6M. Unlike MPI-INF-3DHP~\cite{muco3dhp} and Occlusion Person~\cite{occlusion_person}, there are no good matches to estimate the missing features, which leads to diverging results and high MPJPE results. Finally, Grand Theft Auto-Indoor Motion (GTA-IM)~\cite{caoHMP2020} and NBA2k~\cite{zhu_2020_eccv_nba} are both synthetic 3D HPE datasets like Occlusion Person. Their respective game engines are propriety and compel researchers to purchase physical copies in order to access the data.

%% file: src/conclusion.tex
\section{Conclusion}
\label{sec:conclusion}

In this paper we extend the framework of Zhou et al., 2017~\cite{Zhou_2017_ICCV} to be compatible with additional 2D and 3D HPE datasets, specifically FLIC-Full~\cite{flic_full}, LSP-Extended~\cite{lsp_extended} and Occlusion Person~\cite{occlusion_person}. Furthermore, we fully implement usage of MPI-INF-3DHP~\cite{muco3dhp}, which the original method only partially used. Using these datasets, as well as the baseline datasets of MPII~\cite{mpii} and Human3.6M~\cite{h36m_pami, human36m_iccv}, we perform extensive experiments on the topic of using 2D HPE pre-training to improve the performance on downstream 3D pose prediction. We demonstrate that using 2D pre-training allows us to achieve better downstream MPJPE metrics in less time and with superior generalizability than schemes where 3D data is used exclusively. Additionally, we quantify the role that 2D HPE dataset size plays in overall performance by scaling the number of epochs used to train on smaller 2D datasets, and observing that while it can improve performance, it is not a true substitute for a larger number of sample images. Finally, our best results achieve %
under 64.5mm test MPJPE on H36M. %